\documentclass[10pt,twocolumn,letterpaper]{article}

\usepackage{cvpr}
\usepackage{times}
\usepackage{epsfig}
\usepackage{graphicx}
\usepackage{amsmath}
\usepackage{amssymb}

\usepackage{cvpr}
\usepackage{times}
\usepackage{epsfig}
\usepackage{graphicx}
\usepackage{amsmath}
\usepackage{amssymb}
\usepackage{booktabs}
\newcommand{\eat}[1]{}

\usepackage{caption}
\usepackage{multirow}

\usepackage[T1]{fontenc}    % use 8-bit T1 fonts
\usepackage{url}            % simple URL typesetting
\usepackage{booktabs}       % professional-quality tables
\usepackage{amsfonts}       % blackboard math symbols
\usepackage{nicefrac}       % compact symbols for 1/2, etc.
\usepackage{microtype}      % microtypography
\usepackage[noend]{algpseudocode} %
\usepackage{algorithmicx,algorithm} % both are for algorithm
\usepackage{textcomp}
\makeatletter
\def\BState{\State\hskip-\ALG@thistlm}
\makeatother
\algdef{SE}[DOWHILE]{Do}{doWhile}{\algorithmicdo}[1]{\algorithmicwhile\ #1}%

\usepackage[pagebackref=true,breaklinks=true,letterpaper=true,colorlinks,bookmarks=false]{hyperref}

\usepackage[breaklinks=true,bookmarks=false]{hyperref}

\cvprfinalcopy % *** Uncomment this line for the final submission

\def\cvprPaperID{****} % *** Enter the CVPR Paper ID here

\begin{document}
	
	%%%%%%%%% TITLE
	\title{Recurrent Image Captioner: Describing Images with Spatial-Invariant Transformation and Attention Filtering
	}
	
	\author{Hao Liu\\
		UESTC, China\\
		{\tt\small liuhaosater@gmail.com}
		\and
		Yang Yang\\
		UESTC, China\\
		{\tt\small dlyang@gmail.com}
		\and
		Fumin Shen\\
		UESTC, China\\
		{\tt\small fumin.shen@gmail.com}
		\and
		Lixin Duan\\
		UESTC, China\\
		{\tt\small lxduan@gmail.com}
		\and
		Heng Tao Shen\\
		UESTC, China\\
		{\tt \small shenhengtao@hotmail.com}
	}
	
	\maketitle
	%\thispagestyle{empty}
	
	%%%%%%%%% ABSTRACT
	%%%%%%%%% ABSTRACT
	\begin{abstract}
		Along with the prosperity of recurrent neural network in modelling sequential data and the power of attention mechanism in automatically identify salient information, image captioning, a.k.a., image description, has been remarkably advanced in recent years. Nonetheless, most existing paradigms may suffer from the deficiency of invariance to images with different scaling, rotation, etc.; and effective integration of standalone attention to form a holistic end-to-end system. In this paper, we propose a novel image captioning architecture, termed Recurrent Image Captioner (\textbf{RIC}), which allows visual encoder and language decoder to coherently cooperate in a recurrent manner. Specifically, we first equip CNN-based visual encoder with a differentiable layer to enable spatially invariant transformation of visual signals. Moreover, we deploy an attention filter module (differentiable) between encoder and decoder to dynamically determine salient visual parts. We also employ bidirectional LSTM to preprocess sentences for generating better textual representations. Besides, we propose to exploit variational inference to optimize the whole architecture. Extensive experimental results on three benchmark datasets (i.e., Flickr8k, Flickr30k and MS COCO) demonstrate the superiority of our proposed architecture as compared to most of the state-of-the-art methods.
	\end{abstract}
	
	\section{Introduction}
	%background
	Advanced by the rapid development of smart mobile devices, massive storage, fast Internet and prevailing social media sharing platforms, we have witnessed a tremendous explosion of image data on the Web. Visual content understanding has been long studied in literature, ranging from object recognition \cite{he2015deep,liang2015recurrent,dai2016r}, to image classification \cite{he2015deep,krizhevsky2012imagenet}, to visual semantic analysis \cite{yang2014exploiting,yang2015visual,zhang2014start,hong2015learning,bin2016bidirectional}. Recently, the research focus has gradually shifted to the more challenging visual task, \emph{i.e.}, image captioning, which refers to the process of generating meaningful natural language description for image data, towards deep understanding of visual content. Beside the basic recognition/detection of visual participants, image captioning further explores their interrelationships, which inevitably poses more non-trivial challenges on the model design.
	
	%Image caption has long been viewed as a difficult problem mainly due to that caption generate method not only  be powerful enough to solve the computer vision challenges of determining which objects are in an image, but they must also be capable of capturing and expressing their relationships in a natural language. Image caption need the bridge between natural language processing and computer vision, that is to say a meaningful generation of a natural language description about an image requires a level of image understanding that goes well beyond image classification and object detection, which make it have a great potential to many applications, thus also an importance problem. Due to its difficulty and importance, many papers have devoted to this scope.
	Existing image captioning approaches can be roughly categorized into two schemes: 1) template-based approaches~\cite{kulkarni2013babytalk} and 2) neural network based approaches~\cite{karpathy2014deep}. Most of the early attempts focus on the the former family of captioning methods, which discover static visual participants (e.g., objects, scenes) from images first and then fit them into the templates prepared beforehand. Nonetheless, such approaches are prone to ``general'' descriptions while ignoring the specifics, \emph{e.g.}, the location information. During the past few years, because of the overwhelming data modeling power of deep learning, the state-of-the-art captioning performance has been dominated by neural network based approaches~\cite{chen2015mind,donahue2015long,mao2015learning}, which ``mimic'' machine translation to transform images (CNN-based encoder) into sentences (LSTM-based decoder) in an end-to-end system. One limitation of such paradigms is that they heavily depend on the entire visual representations of images, thereby neglecting the fine-grained details due to the complexity and arbitrariness of image content.
	
	Intuitively, when watching a piece of image, one might only attend to a small proportion of objects among various participants, \emph{e.g.}, the leading character among all the actors in a single screen. Inspired by this observation, attention mechanism~\cite{xu2015show,mnih2014recurrent} has been introduced to facilitate the encoder-decoder framework to explore more detailed aspects. By simulating human visual perception system, attention tries to identify certain particular salient parts of a given image and has successfully facilitated a variant of applications, such as image generation~\cite{gregor2015draw} and object detection~\cite{zhou2015learning}. In~\cite{xu2015show}, the description generating process condition on the hidden states images encoder, rather than on one single context vector only, that is to selectively attend to parts of the scene while ignoring others.
	
	It is known that a commonly-used choice of visual encoder is traditional Convolutional Neural Network (CNN), which provides limited support for exploring spatial invariant property in input images~\cite{jaderberg2015spatial}. Hence, this drawback may cause the encoder-decoder framework vulnerable to images with large variance, such as scaling, rotation, translation, etc. Besides, existing attention mechanism normally serves as a standalone component (\emph{e.g.}, select feature maps from a fixed some low levels map of a pretrained CNN layers to represent images~\cite{xu2015show}) attached to certain encoder-decoder framework, which cannot be formulated as a holistic end-to-end system for coherently unifying the processes of visual encoding, attending to salient parts and decoding to sentences.

	In this paper, we propose a novel architecture, termed \emph{Recurrent Image Captioner} (RIC), which unifies spatially invariant property, automatical attention filtering mechanism together with a recurrent feedback loop between CNN-based encoder and LSTM-based decoder to form an end-to-end formulation for image captioning task. Specifically, inspired by~\cite{jaderberg2015spatial}, we first equip CNN-based visual encoder with a differentiable layer to enable spatially invariant transformation of visual signals. Then, in order to achieve indepth integration of attention mechanism, we add a differentiable attention filter to automatically capture the semantically vital regions in a dynamic manner based on previous visual representations and generated captions. Besides, due to the sequential nature of image description and generation process, we further introduce a loop from decoder to spatial transformation layer to feedback the invariant information in a recurrent way. The recurrent feedback loop helps to gradually bridge the semantic gap between low-level visual features and high-level caption semantics, thereby generating more accurately captions by translating the spatially invariant visual signals filtered by the attention module. For optimization, there are plenty of sequential attention models trained with reinforcement learning techniques, such as policy gradients~\cite{mnih2014recurrent}. It is worth noting that both of the aforementioned components are differentiable, which makes our proposed RIC architecture easy to optimize with standard backpropagation. We propose to exploit variational inference~\cite{frigola2014variational} to optimize the whole architecture.

	We summarize our contributions as follows:
	\begin{itemize}
		\item We propose a novel image captioning architecture, which formulates the processes of visual encoding, spatial transformation, attention filtering as well as sentence decoding in a recurrent feedback loop.
		\item Both the spatial transformation layer and attention filter are differentiable, which makes the whole structure easy to optimize with the assistance of variational inference.
		\item Extensive experiments on Flickr8k, Flickr30k and MS COCO datasets demonstrate the superiority of our proposal as compared to the state-of-the-arts.
		%\item We introduce an efficient framework, HAM, a recurrent hierarchical attention-based image caption generators that extend currently attention based encoder-decoder methods by construct a hierarchical module to learn to firstly extract spatial invariant information and secondly  place higher attention on them.
		%\item We quantitatively validate the usefulness of hierarchical attention model in caption generation with state of the art performance on three benchmark datasets: Flickr8k, Flickr30k and the MS COCO dataset,  by which we firstly show its benefit and ways of taking spatial invariant information int consideration for image caption generation.
		%\item Experimentally we show that by incorporating our hierarchical attention module, we are able to directly learn from the raw image input to generate accurate descriptions.
	\end{itemize}
	
	The rest of this paper is organized as below. Section 2 briefly reviews related work on image captioning. Section 3 elaborated the proposed Recurrent Image Captioner. In Section 4, we report experiments on three image benchmarks, followed by the conclusion in Section 5.
	
	% \begin{figure}[t]
	% \begin{center}
	%    \includegraphics[width=0.8\linewidth]{latent_GMM.png}
	% \end{center}
	%    \caption{Example of caption.  It is set in Roman so that mathematics
	%    (always set in Roman: $B \sin A = A \sin B$) may be included without an
	%    ugly clash.}
	% \label{fig:long}
	% \label{fig:onecol}
	% \end{figure}

	\begin{figure*}[t]
		\begin{center}
			\includegraphics[width=0.8\linewidth]{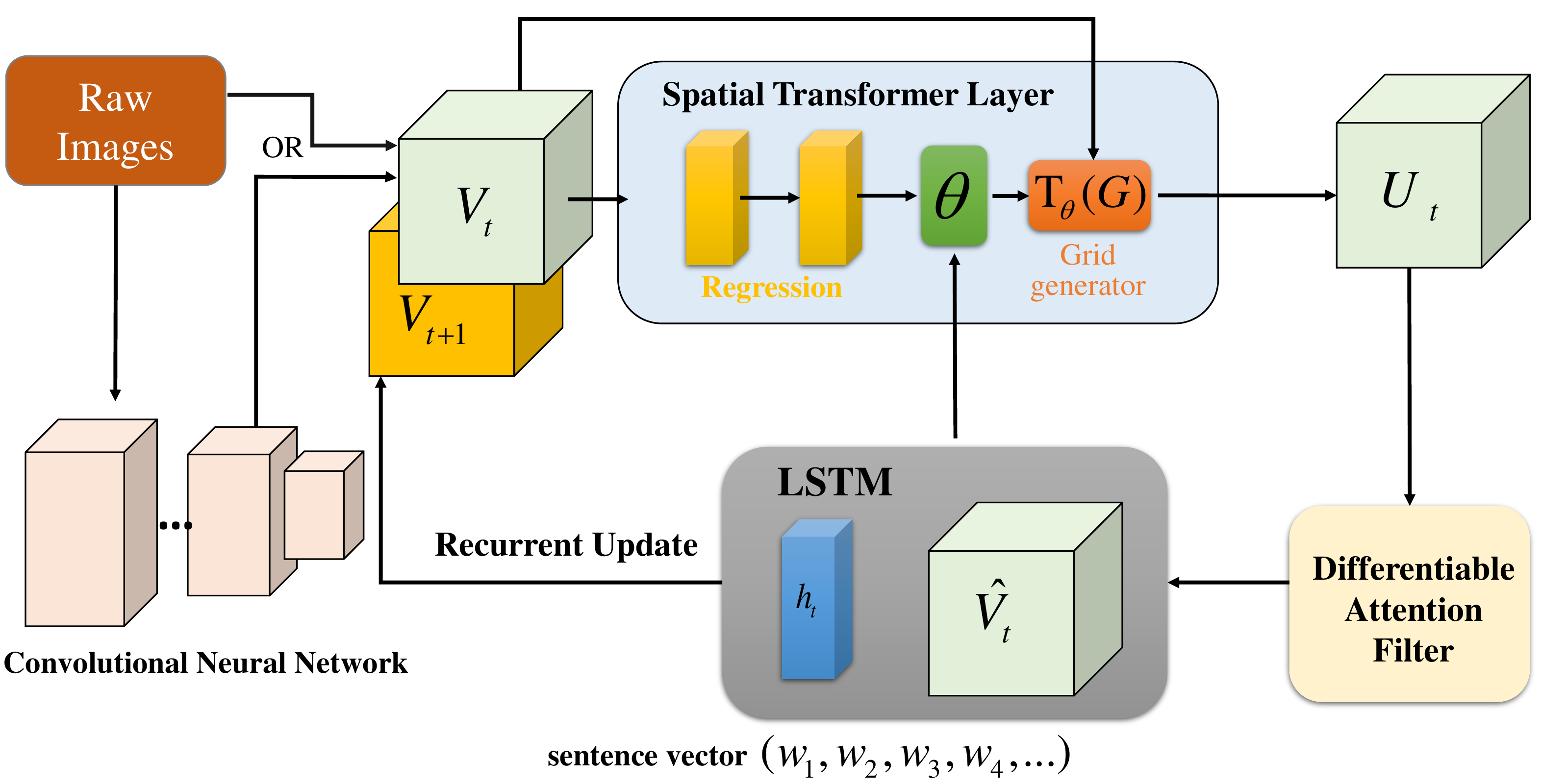}
		\end{center}
		\caption{The overall flowchart of the proposed Recurrent Image Captioner architecture.
			%The recurrent neural network used in updating memory:
			%The architecture of our main architecture: At time step t-1, the input feature map $V_{t-1}$ is passed to a spatial attention module which regresses the transformation parameters $\theta$. Then the regular spatial grid G over $V_{t-1}$ is transformed to the sampling grid $T_\theta (G)$, producing the warped output feature map $V_t$ .
		}
		\label{fig:main}
	\end{figure*}
	
	\section{Related Work}
	In this section, we briefly review related work on image captioning and attention mechanism. Generally, existing captioning can be roughly categorized into two families: template-based approaches ~\cite{elliott2013image, kuznetsova2012collective} and neural network based approaches~\cite{karpathy2015deep,karpathy2014deep,xu2015show}. Most of the early attempts focus on the the former family of captioning methods, which discover static visual participants (e.g., objects, scenes) from images first and then fit them into the templates prepared beforehand. For instance, in \cite{farhadi2010every, kulkarni2011baby}, semantic concepts are detected and then fed into templates to construct sentences. In~\ \cite{li2011composing} concepts are first discovered and then directly combined together. Nonetheless, such approaches are prone to ``general'' descriptions while ignoring the specifics, \emph{e.g.}, the location information.
	
	Recently, various neural network based methods have been proposed for generating image descriptions. The very first approach to use neural network to generating image caption generation neural work was~\cite{kiros2014multimodal}, which
	introduces a multimodal log-bilinear model biased by features from the image. This work was then followed by \cite{kiros2014unifying} in which an explicit way of ranking and generation was introduced. There are also recurrent neural network based approaches to image caption generation, such as \cite{bahdanau2014neural, cho2014learning,donahue2015long,sutskever2014sequence,vinyals2015show}, where a commonly used RNN structure is a LSTM. They represent images as a single feature vector from the top layer of a pre-trained convolutional network.
	\cite{mao2014deep} proposes to learn a joint embedding space of both images and descriptions for ranking and caption generation. The model learns to score sentence and image similarity as a function of R-CNN object detections with outputs of a bidirectional RNN. \cite{fang2015captions} proposes a three-step pipeline for generation by incorporating object detections. Recently, we have witnessed an increasing trend of incorporating attention mechanism~\cite{fang2015captions,karpathy2015deep,kiros2014unifying,vinyals2015show,xu2015show} into neural networks for boosting computer vision and artificial intelligence tasks, such as object recognition, image caption, etc. There are also attention based work to handle semantic related tasks, such as~\cite{fang2015captions}.
	
	In comparison to existing approaches, it is important to note that our proposed architecture introduce two differentiable modules, which enables our method capturing global information in images and learning spatially invariant structure in a holistic end-to-end system. Furthermore, we show the benefit of our architecture by defining a variational loss function which employs an autoencoder to regularize the process of image captioning.
	
	\section{Recurrent Image Captioner}
	In this section, we elaborate the details of the proposed Recurrent Image Captioner, including architecture overview, spatial transformation layer, differential attention filter as well as loss functions.
	
	\subsection{Architecture Overview}
	The overall flowchart of our proposed RIC is illustrated in Fig.~\ref{fig:main}, which comprises four major components: (a) basic CNN encoder, (b) spatial transformation layer, (c) differentiable attention filter and (d) language decoders. Given a set of training images, we feed them into the CNN encoder, which outputs visual latent codes, denoted as $V_{t}$. Then, the spatial transformation layers converts the visual latent codes into spatially invariant signals, denoted as $U_{t}$, which are subsequently passed to the attention filter to distill the most semantically informative parts. The conventional LSTM is then employed to decode the filtered signals to sentences. After after a duration of steps, the decoding outputs are successively added to the distribution that generates the captions rather than emitting a word in a single step. This shows that our method capture more higher concept instead of simple word-level information in image. Finally, we recurrently feedback the output parameters of the decoder for updating the visual latent codes and the transformation parameter $\theta$.
	
	%To give a overview of our architecture, the bfasic of HAM network is similar to that of other attentive caption model, both consist of an encoder network which take images as input, output latent codes and an decoder network which take samples of latent code as input and use them to learn a conditional distribution of captions.
	%However there are four key differences.
	%Firstly, in addition to use convolutional neural network as encoder, we also propose a recurrent selective module which can directly generate captions condition on raw images input. This is not only a novel architecture which generate accurate caption directly on raw images input, but also coherent with the sequence nature of language, we will show this benefit in experiment section.
	%Secondly,  while previous work using CNN to extract abstract code from images, e.g.  selectively choose features by condition on generated captions, our method propose to extract spatial invariant from images in a current way, as you can see in Fig.~\ref{fig:main}.
	%Thirdly,  the decoder’s outputs are successively added to the distribution that generate the caption after a duration of steps , as opposed to emitting a word in a single step. This shows that our method capture more higher concept instead of simple word-level information in image.
	%And fourthly, in our architecture, the attention mechanism is a hierarchical dynamically updated mechanism, used for the encoder and decoder to modify its hierarchical representation.
	
	\subsubsection{Basic CNN Encoder}
	Following conventions, we use CNN as encoder as well. Specifically, we employ VGG-16~\cite{simonyan2014very} to extract a set of visual feature vectors for a given image, denoted as
	\begin{equation}
	\mathcal{A}  = \{ {a _1},{a _2}, \cdots ,{a _L}\},
	\end{equation}
	where each ${a_i} \in {\mathbb{R}^d}$ corresponds to a part of the image, $L$ is the number of image parts and $d$ is the dimensionality of the feature space.
	
	%decoder
	\subsubsection{Basic LSTM Decoder}
	
	%\begin{figure}
	%\begin{center}
	%   \includegraphics[width=0.8\linewidth]{lstm.PNG}
	%\end{center}
	%   \caption{\textbf{This figure need revisited} LSTM cell, lines with bolded squares imply projections with a learnt weight vector. Each cell learns how to weigh its input components (input gate), while learning how to modulate that contribution to the memory (input modulator). It also learns weights which erase the memory cell (forget gate), and weights which control how this memory should be emitted (output gate).cited from \cite{xu2015show} }
	%\label{fig.LSTM}
	%\end{figure}
	
\begin{figure}[h]
		\begin{center}
			\includegraphics[width=\linewidth]{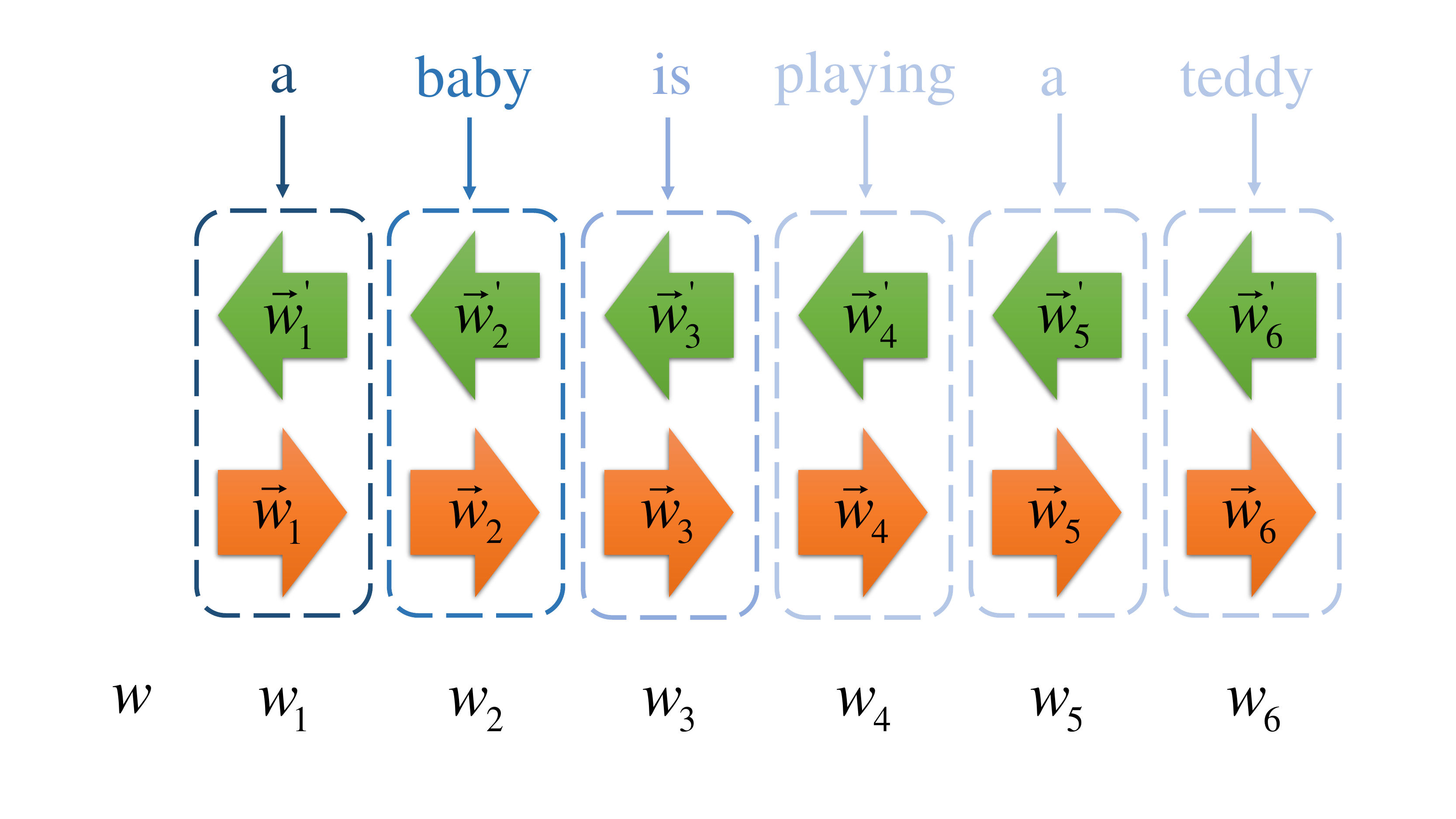}
		\end{center}
		\caption{Illustration of Bidirectional LSTM for encoding captions.}
		\label{fig:lan}
	\end{figure}

	For better decoding, we propose to preprocess image captions with Bidirectional LSTM~\cite{}. In a Bidirectional LSTM model, the two LSTMs~\cite{hochreiter1997long} jointly process the input caption sequence from both forward and backward directions. The forward LSTM computes the sequence of forward hidden states, denoted as $\{\vec{w}_1,\vec{w}_2,...,\vec{w}_M\}$, whereas the backward LSTM computes the sequence of backward hidden states $\{\vec{w'}_1,\vec{w'}_2,...,\vec{w'}_M\}$, where $M$ is the number of words in the caption. These hidden states are then concatenated together into the sequence $\mathcal{W} = [w_ 1, w_2,...,w_M ]$, with $w_i = [\vec{w}_i, \vec{w'}_i], 1 \leq i \leq M$. The process is illustrated in Fig.~\ref{fig:lan} for a graphical interpretation. It is worth noting that instead of representing caption with 1-of-$K$ encoded words, we can regard the bidirectional LSTM as a generator of word embedding, which is capable of capturing semantical relationships in language.
	
	%We adopt a common approach to represent word as vectors ${\rm{y}} = \{ {y_1},{y_2}, \cdots ,{y_C}\} ,{y_i} \in {R^k}.$ where K is the size of the vocabulary and C is the length of the caption. Let y be the input caption, represented as a sequence of 1-of-K encoded words y = $(y_1, y_2,..., y_N)$, where K is the size of the vocabulary and N is the length of the sequence. We obtain the caption sentence representation by first transforming each word $y_i$ to an m-dimensional vector representation $h_i$ ,i = 1,...,N using the Bidirectional RNN. In a Bidirectional RNN, the two LSTMs \cite{hochreiter1997long} with forget gates \cite{gers2000learning} process the input sequence from both forward and backward directions. The Forward LSTM computes the sequence of forward hidden states $[\vec{y}_1,\vec{y}_2,...,\vec{y}_N]$ , whereas the Backward LSTM computes the sequence of backward hidden states $[\vec{y'}_1,\vec{y'}_2,...,\vec{y'}_N]$. These hidden states are then concatenated together into the sequence $y = [y_ 1, y_2,...,y_N ]$, with $y_i = [\vec{y}_i, \vec{y'}_i]; 1 ≤ i ≤ N$. See Fig. ~\ref{fig:lan} for a graphical interpretation. As in \cite{xu2015show}, they propose to extract features from a lower convolutional layer unlike previous work which instead used a fully connected layer in order to obtain a correspondence between the feature vectors and portions of the 2-D image.
	
	To model the decoding process, we exploit Long Short-Term Memory architecture~\cite{hochreiter1997long} with extension of forget gates~\cite{gers2000learning} due to its excellent capability of modelling long-range dependencies in real sequential data. We follow the implementation of LSTM as in~\cite{zaremba2014recurrent}.

	\subsection{Spatial Transformation Layer}\label{sec:spatial invariance}
	In this part, we present a spatial transformation layer extended from~\cite{}, which is installed on top of the CNN encoder to add spatially invariant information into visual signals. As shown in Fig.~\ref{fig:main}, different from the original model as in~\cite{}, the transformation parameter $\theta$ in our spatial transformation layer not only depends on the internal regression network but also conditions on the LSTM module in the decoder, which forms the foundation of the recurrent updating loop between encoder and decoder.
	
	%This section will present one of our main contribution, spatial top-down attention model. In loss function part, we will show how to incorporate it with different encoder attention to learn a hierarchical representation.
	%In order to extract spatial invariant information from images, we propose a spatial top-down attention module to extract invariant information and attach higher attention to them.
	
	%Specifically, We use a hierarchical attention model driven by some technologies from computer graphics and multilayer neural network to jointly model the context vector ${\hat z_t}$.
	
	Suppose we are given the input feature cube $V_{t}\in {\mathbb{R}^{H \times W \times C}}$ at time step $t$, where $H$, $W$ and $C$ represents height, width and the number of channels, respectively, then we calculate the transformation parameter $\theta$ as follows:
	\begin{equation}\label{eq:theta}
	\theta=f_{recurrent}(V_{t},h_{t},\hat{V}_{t}),
	\end{equation}
	where $\hat{V}_{t}$ denotes the output of the attention filter in the $t$-th updating loop, which will be elaborated in the next subsection. $h_{t}$ is the hidden states of the LSTM-based decoder in the $t$-th updating loop. $f_{recurrent}(\cdot)$ is the recurrent updating function of the transformation parameter $\theta$. $f_{recurrent}(\cdot)$ can be implemented as a multilayer perception or as a gated activation function, such as LSTM or Gated Recurrent Unit (GRU)~\cite{cho2014learning}. Both LSTM and GRU have been applied to learn long-term dependencies. In our experiments, we have found that GRU provides slightly better performance than LSTM, therefore we choose to use GRU as the implementation of $f_{recurrent}(\cdot)$.
	
	%$U = V_0 \in {R^{H \times W \times C}}$ is the input feature cube. We hypothesis that by using a transform on the grid is benefit to attention.
	%This network takes the input feature map $U \in {R^{H \times W \times C}}$  with width W, height H and C channels and output $\theta$, the parameters of the transformation $T _\theta$ to be applied to feature map. The size of $\theta$ can vary depending on the transfomation type that is parameterised.
	%$\theta  = {f_{att}}({h_{t - 1}},{\hat V_t})$
	%where $\hat{V_t}$ is the output of selective attention model, we will discuss it in Sec.~\ref{Sec.sele} ${f _{att}}$ is transition function can be implemented as a multilayer perception or as a gated activation function, such as Long Short-Term Memory (LSTM)\cite{hochreiter1997long} or a Gated Recurrent Unit (GRU) \cite{cho2014learning}. Both LSTM and GRU have been proposed to address the issue of learning long-term dependencies. In experiments we have found that GRU provides slightly better performance than LSTM , and therefore the GRU is used.
	% For now we assume that $\alpha _i,i=1,...,L$ are annotations vectors as in \cite{xu2015show}, as we will detail in Sec.~\ref{Sec.sele}, $\alpha _i, i=1,...,L$ can be row vectors from a matrix $V_0$ which is a output of selective attention when input are raw image.
	% Note that $f_{att}$ can take any form, such as a fully-connected network or a convolutional network, but should include a final regression layer to produce the transformation parameters $\theta$.
	
	Then we assemble the transformation, which should be differentiable, as $T_{\theta}=P_\theta B$, where $P_\theta$ is a matrix parameterised by $\theta$ and $B$ is the representation of a grid. As pointed out in~\cite{}, we may possibly learn to predict $\theta$, and also learn $B$ for the task.
	
	After the transformation, we apply a learnable kernel as a mask. Here, we refer pixel to a particular location, which is not necessarily a real visual pixel.
	\begin{equation}
	\begin{array}{l}
	\hat{U}_{i,t} = \sum\limits_m^{H^V_t} {\sum\limits_n^{W^V_t} {V_{nm,t}^c} } k(x_i - m|{\Phi _x})k(y_i - n|{\Phi _y}),\\
	\forall i \in [1, \cdots ,H^U_ t W^U_t], \forall c \in [1, \cdots ,C]
	\end{array}
	\end{equation}
	where each $(x _i, y_i)$ coordinate in $T_\theta $ defines the spatial location in the input where a kernel is applied to get the value at a particular pixel in the output $V_{i,t}^c$. $H_t^V$  and $W_t^V$ are the dimensions of $V^c_{nm,t}$ , while $H_t^U$ and $W_t^U$ are dimensions of $\hat{U}_{i,t}$. $\Phi _x$ and $\Phi _y$ are the parameters of a generic sampling kernel $k(\cdot|\cdot )$, which is differentiable. To preserve spatial consistence we apply identical sampling/transformation strategy to each channel of the input. Mask has been used in many image modeling work, such as~\cite{dinh2014nice,dinh2016density,larochelle2011neural,mathieu2015masked,van2016pixel}. Different from previous work, our model do not use pre-trained mask but learn a proper mask using a recurrent neural network.
	
	\subsection{Differentiable Attention Filter}\label{sec:attention_filter}
	In this part, we introduce the attention filtering module which serves as an important role in our architecture. Specifically, we take inspiration from the differentiable attention mechanisms for handwriting synthesis~\cite{graves2013generating} and Neural Turing Machines~\cite{graves2014neural,gregor2015draw}. We use an array of 2-D Gaussian filters, which is an explicit 2-D form of attention map. The output of this module varies smoothly on different locations.
	The real-valued grid center $(g_X, g_Y)$ and stride $\delta$ determine the mean location $(\mu _X^i, \mu _Y^i)$ of the filter at row $i$, column $j$ in the patch as follows:
	\begin{equation}
	\left\{
	\begin{array}{c}
	\mu _X^i = {g_X} + (i - N/2 - 0.5)\delta,\\
	\mu _Y^j = {g_Y} + (j - N/2 - 0.5)\delta,
	\end{array}
	\right.
	\end{equation}
	where $N$ is the dimensionality of input.
	The isotropic variance $\sigma^2$ of the Gaussian filters, and a scalar intensity $\gamma$ that multiplies the filter response are defined as follows:
	\begin{equation}
	\left\{
	\begin{aligned}
	&(\hat{g}_Y ,\hat{g}_Y,\log {\sigma ^2},\log \hat{\delta},\log \gamma ) = Ph_t,\\
	&{g_X} = \frac{{H_t^V + 1}}{2}(\hat{g}_X  + 1),\\
	&{g_Y} = \frac{{W_t^V + 1}}{2}(\hat{g}_Y  + 1),\\
	&\delta  = \frac{{\max (H_t^V,W_t^V) - 1}}{{N - 1}}\hat{\delta},
	\end{aligned}
	\right.
	\end{equation}
	where $P$ is a linear mapping matrix. Given an input image or feature map of size $N \times N$, where the variance, stride and intensity are emitted in the log-scale to ensure positivity.
	
	The horizontal and vertical filter bank matrices $F_X$ and $F_Y$ are defined as follows:
	\begin{equation}
	\left\{
	\begin{array}{l}
	{F_X}[i,m] = \frac{1}{{{Z_X}}}\exp ( - \frac{{{{(m - \mu _X^i)}^2}}}{{2{\sigma ^2}}}),\\
	{F_Y}[j,n] = \frac{1}{{{Z_Y}}}\exp ( - \frac{{{{(n - \mu _Y^j)}^2}}}{{2{\sigma ^2}}}),
	\end{array}
	\right.
	\end{equation}
	where $(i,j)$ is a point in the attention patch, $(m,n)$ is a point in the input image, and $Z_X, Z_Y$ are normalization constants that ensure $\Sigma _m F_X[i, m]$ = 1 and $\Sigma _n F_Y [j, n]$ = 1. We can view it as the process of producing two arrays of 1D Gaussian filter banks, whose filter locations and scales are computed from the LSTM hidden states. Given the above equation, we then apply this attention filter on top of spatial transformation layer.
	
	One way is $\hat{V_t}$ := $\gamma F_X U_t F_Y^T$ given the previous layer output $U_t$. Another way is to calculate the convolution of the filter map $F_X$, $F_Y$ and $U_t$, followed by a pooling operation. The output of this module will serve as latent code in our variational loss function:
	\begin{equation}
	\left\{
	\begin{aligned}
	&{{\tilde V}_{1}^c} = {U^c_t}*{F_X},~c=1,2,\ldots,C;\\
	&{{\tilde V}_{2}^c} = {{\tilde V}_{1}^c}*{F_Y},~c=1,2,\ldots,C;\\
	&{\hat{V}_{t}^c}\leftarrow \mathbf{pool}({{\tilde V}^c_{2}}),~c=1,2,\ldots,C.
	\end{aligned}
	\right.
	\end{equation}
	Here $*$ is convolution operation performed at every channel, $\mathbf{pool}(\cdot)$ is the pooling operation, ${{\tilde V}_{1}^c}$ and ${{\tilde V}_{2}^c}$ are both intermediate outputs. Concretely, we sequentially use $F_X$ and $F_Y$ to , after this, we apply a pooling to the output to get $\hat{V}^t$.
	The output of the differentiable attention filter is fed into the LSTM module for sentence generation. In the following part, we will define two loss functions based on both of the aforementioned filtering mechanisms.
	
	%There are similar approaches in computer graphics-based autoencoders \cite{tieleman2014optimizing} such as affine transformations.

	% \textbf{Gaussian Selective Attention}
	% \[\begin{array}{l}
	% P({z_t},{z_{1:t - 1}}) = N(\mu ({h_{t - 1}}),\sigma ({h_{t - 1}}))\\
	% \mu ({h_{t - 1}}) = \tanh ({W_\mu }{h_{t - 1}})\\
	% \sigma ({h_{t - 1}}) = \exp (\tanh ({W_\sigma }{h_{t - 1}}))
	% \end{array}\]
	% where ${W_\mu } \in {R^{D \times n}},{W_\sigma } \in {R^{D \times n}}$ are the learned model parameters, and n is the dimensionality of $h_t$ , the hidden state of the generative LSTM. Similar to \cite{bachman2015data} , we have observed that the model performance is improved by including dependencies between latent
	% variables.
	% While Gaussian Selective Attention can be used in variational learning framework for learning better latent representation , experimentally, we found that when paired with selective attention model, spatial top-down attention mechanism work better.

	\subsection{Loss Functions}\label{sec:loss}
	In this subsection, we devise two loss functions for optimization of the RIC architecture.
	
	\subsubsection{Discriminative Attention Loss}
	Different from attention-based image captioning~\cite{xu2015show}, we first compute the weights of the visual signals as follows:
	\begin{equation}
	\alpha _i^t = \frac{{\exp ({v^T}\tanh (RV_t^c + Q{V_{i,t}} + L{w_{pre}} + b))}}{{\sum\nolimits_{k} {\exp ({v^T}\tanh (RV_{k,t}^c + Q{V_{k,t}} + L{w_{pre}} + b))} }}
	\end{equation}
	where $w_{pre}$ is the generated caption in the previous step. %$y_i, i=1,2,...,L$ are vectors we computed in previous section.
	$v, L, R, Q$ and $b$ are variables to be learned. Then we define $P({{w_t}} |V_{t-1},{w_{1\sim t - 1}}) \propto \exp (E{w_{t - 1}} + {L_h}{h_t} + F{\hat z_t})$ as the probability of the $t$-th vectors given the previous generated $t-1$ vectors, where $E$, $L_h$ and $F$ are learnable matrix. ${\hat z_t}$ is the context vector. The $w_t$ with the highest value is then selected.
	
	We further define $s_{t,i}$ as an indicator one-hot variable which is set to $1$ if the $i$-{th} location is the one used to extract visual features. By treating the attention locations as intermediate latent variables, we can assign a multinoulli distribution parametrized by $\{\alpha _i\}$ , and view it as a random variable, similar to \cite{xu2015show}:
	\begin{equation}
	\left\{
	\begin{aligned}
	&p({s_{t,i}} = 1|{s_{j < t}},a) = {\alpha^t_{k}},\\
	&\mathop {{\hat z_t}}   = \sum\nolimits_i {{s_{t,i}}{a_i}}.
	\end{aligned}
	\right.
	\end{equation}
	Given this formulation of context vector, we now employ the following loss function, which is comprised of a sequence output probabilities for measuring the accuracy of captions:
	\begin{equation}
	\label{eq.objective}
	\mathcal{L}_s = \sum\limits_s {P(\left. s \right|V_0)} \log P(\left. y \right|s,V_0).
	\end{equation}
	where $s$ denotes the set of $s_{t,i}$ variables. $V_0$ is the initial feature cube.
	%$P(\left. {{y_t}} \right|V,{y_{1\sim t - 1}}) \propto \exp ({L_0}(E{y_{t - 1}} + {L_h}{h_t} + F{\hat z_t}))$
	
	Our proposed architecture is also natural to incorporate with discriminative loss~\cite{fang2015captions}, which benefits training with the discriminative supervision. Let $r_j$ be the score of word $j$ after the max pooling layer, and $\Omega$ be the set of all words that occur in the caption $w$.  We define the discrminative loss as follows:
	\begin{equation}
	{\mathcal{L}_d} = \frac{1}{Z}\sum\limits_{j \in \Omega} {\sum\limits_{i \ne j} {\max (0,1 - ({r_j} - {r_i}))} },
	\end{equation}
	where $Z$ is the normalizer. Our goal is to minimize this loss function $L= L_s + \lambda L_d$, where $\lambda$ is a constant weight factor.
	%\begin{equation}
	%\begin{array}{l}
	%\min \frac{{\partial {L}}}{{\partial w}} = \\
	%\sum\limits_s {P(\left. s \right|U)} [\frac{{\partial \log P(\left. y \right|s,U)}}{{\partial w}} + \log P(\left. y \right|s,U)\frac{{\partial \log P(\left. s \right|U)}}{{\partial w}}]\\
	%+ \lambda  \frac{{\partial {L_d}}}{{\partial \omega }}
	%\end{array},
	%\end{equation}
	We adopt adaptive stochastic gradient descent to train the model in an end-to-end manner. The loss of a training batch is averaged over all instances in the batch. In experiment we show that by using this loss function with the proposed architecture, our method outperforms most state-of-the-art methods.
	
	\subsubsection{Variational Autoencoder Regularized Loss}
	To further improve our proposed architecture, we define a novel variational learning based loss function. When using this sort of loss function, the architecture takes the raw images instead of VGG features as input. In the following recurrent process, it works like an autoencoder, and serves as a regularization for image captioning. Variational autoencoder for image caption generation is also explored in recent work~\cite{pu2016variational}. Apart from this work, which merely encodes images to latent codes, our encoding-decoding process is a recurrent process that not only depends on visual information but also language sentences.
	Specifically, we first generate latent codes from attention filter as follows:
	\begin{equation}
	P({Z_1}) = \mathcal{N}({\mu _\phi }({\hat{V}_{1}}),diag(\sigma _\phi ^2({\hat{V}_{1}}))),
	\end{equation}
	where $\mu _ \phi $ and $\sigma _\phi $ are parameters of Gaussian distribution. Then we conduct the following calculation
	\begin{equation}
	\left\{
	\begin{aligned}
	&P({Z_t}|{Z_{1:t - 1}}) = \mathcal{N}(\mu ({h_{t - 1}}),\sigma ({h_{t - 1}})),\\
	&\mu ({h_{t - 1}}) = \tanh ({W_\mu }{h_{t - 1}}),\\
	&\sigma ({h_{t - 1}}) = \exp (\tanh ({W_\sigma }{h_{t - 1}})),\\
	&{z_t}\sim P({Z_t}|{Z_{1:t - 1}}) = N(\mu ({h_{t - 1}}),\sigma ({h_{t - 1}})),\\
	&{h_t} = LSTM({h_{t - 1}},[{z_t},[{h_{t - 1}},{w_{t - 1}}]]),\\
	%{c_t} = {c_{t - 1}} + decode({h_t}),\\
	&V_0\sim P( \cdot |w,{Z_{1:T}}) = \prod\limits_i {P( \cdot |w,{Z_{1:T}})}.
	\end{aligned}\right.
	\end{equation}
	where $Z_t$ denotes encoded latent code at time step $t$. $W_\mu$ and $W_\sigma$ are learnable matrix. $[ \cdot,\cdot ]$ means concatenation operation. $P( \cdot |y,{Z_{1:T}})$ is chosen to be Bernoulli distribution. We note that some similar work also try to decode the encoded image code~\cite{gregor2015draw,mansimov2015generating}. However, they just generate image from caption while we utilize this kind of generative process as regularization.
	
	Denote the above decoder as $q_\lambda$, i.e., inference network in variational autoencoder, our goal is to maximize the following variational lower bound:
	\begin{equation}
	\begin{aligned}
	&{E_{{q_\phi }(Z|{V_0},w)}}[\log {p_\varphi }(w|Z)]  \\
	&+\beta \{ \sum\nolimits_Z {{q_\phi }(Z|{V_0},w)\log {q_\lambda }(V_0|w,Z)} \\
	&- {D_{KL}}({q_\phi }(Z|{V_0},w)||{q_\lambda }(Z|w))\}
	\end{aligned}
	\end{equation}
	where $\beta$ is a balance parameter. $q_\phi (Z|V_0,w)$ is the recurrent generative network consists of spatial transformation layer and differentiable attention filter. $p_\varphi (w|Z)$ is the likelihood function of generating caption.
	%In view of using autoencoder on images as regularization for caption generation,  we note that in a recently work %\cite{pu2016variational} the authors propose to use image auto-encoder which use solely images as regularization, however, our %architecture is natural to incorporate sentences information in generative network, which proved to be benefit to image caption.
	If we set $\beta$ to zero, this loss function is exactly the same as that in~\cite{karpathy2014deep,vinyals2015show,wu2015value,xu2015show}. This shows that our framework is more general. In all of our experiment, we use the ADAM optimizer proposed in~\cite{kingma2014adam}.
	
	\section{Experiment}
	\label{sec.exp}
	In this section, we will evaluate our method on several benchmarks,  we also compare three different implementation of our architecture:  solely use spatial transformer layer, use both spatial transformer layer and differentiable attention filter, and with variational autoencoder.
	\begin{table*}[]
		\centering
		
		\caption{Results of  BLEU \cite{papineni2002bleu} and METEROR \cite{banerjee2005meteor} on the Flickr 8k \cite{hodosh2013framing}  dataset. }
		\label{8k}
		\begin{tabular}{@{}llllllll@{}}
			\toprule
			Methods         &  BLEU-1   & BLEU-2   & BLEU-3  & BLEU-4 & PPL & METEROR\\ \midrule
			VggNet+RNN              & 0.562 & 0.375  & 0.245 & 0.166 &  15.71 & - &  \\
			Log Bilinear\cite{kiros2014multimodal}    & 0.663 &  0.425 & 0.275 & 0.173 & - & 17.31 &\\
			GoogLeNet+RNN           & 0.565 & 0.385 & 0.277 & 0.163 & - &  15.71 & \\
			Hard-Attention \cite{xu2015show}         & 0.675 & 0.464 & 0.313 & 0.212 & - & 20.30 &  \\
			Joint model with ImageNet \cite{pu2016variational}            & 0.70 & 0.49 & 0.33 & 0.22 & - & 15.24 &\\
			Attributes-CNN+LSTM \cite{wu2015value} &  0.74 & 0.54 & 0.38& 0.27 & 12.60 & -  \\
			RIC with STL & 0.687 & 0.478 & 0.331 & 0.220 & 15.02 & 20.54 & \\
			RIC with both STL and DAF   & 0.696 & 0.481 & 0.326 & 0.225 & 15.11 & 22.73  & \\
			RIC with variational autoencoder   & 0.723 & 0.524 & 0.353 & 0.217 & 15.71 & 23.12  & \\
			Human\cite{chen2014learning}  & - & - & - & - & - & 25.5 &\\  \bottomrule
		\end{tabular}
	\end{table*}

	\begin{table*}[]
		\centering
		\caption{Results of  BLEU \cite{papineni2002bleu} and METEROR \cite{banerjee2005meteor} on the Flickr 30k \cite{yang2011corpus}  dataset. }
		\label{30k}
		\begin{tabular}{@{}lllllll@{}}
			\toprule
			Methods         &  BLEU-1   & BLEU-2   & BLEU-3  & BLEU-4 & PPL & METEROR\\ \midrule
			VggNet+RNN            & 0.591 & 0.382 & 0.254 & 0.173 & 18.83 & - \\
			Log Bilinear\cite{kiros2014multimodal}     & 0.601 & 0.381 & 0.257 & 0.174 & - & 16.88 \\
			GoogLeNet+RNN           & 0.585 & 0.396 & 0.263 & 0.171 & 18.77 & - \\
			Hard-Attention \cite{xu2015show}         &  0.674 & 0.445 & 0.307 & 0.206 & - &18.46    \\
			semantic attention \cite{you2016image} &0.647& 0.460 &0.324 &0.230 &- &0.189 \\
			Joint model with ImageNet \cite{pu2016variational}           & 0.69 & 0.50 & 0.35 & 0.22& 16.17 & - \\
			Attributes-CNN+LSTM \cite{wu2015value} & 0.73 & 0.55 & 0.40& 0.28 & 15.96 & -  \\
			RIC with STL  & 0.681 & 0.489 & 0.338 & 0.223 & 15.67 & 18.77 \\
			RIC with STL and DAF & 0.684 & 0.513 & 0.352 & 0.233 & 15.77 & 19.87 \\
			RIC with variational autoencoder   & 0.745 & 0.528 & 0.375 & 0.244 & 15.94 & 20.16 \\
			Human\cite{chen2014learning}   & - & - & - & -& -& 22.9 \\ \bottomrule
		\end{tabular}
	\end{table*}

	\begin{table*}[]
		\centering
		\caption{Results of  BLEU \cite{papineni2002bleu} and METEROR \cite{banerjee2005meteor} on the MSCOCO \cite{chen2015microsoft} dataset.
		}
		\label{MSCOCO}
		\begin{tabular}{@{}lllllll@{}}
			\toprule
			Methods         &  BLEU-1   & BLEU-2   & BLEU-3  & BLEU-4 & PPL & METEROR\\ \midrule
			VggNet+RNN     & 0.612 & 0.425 & 0.285 & 0.194 & 13.16 & 19.02 \\
			GoogLeNet+RNN           & 0.606 & 0.405 & 0.268 & 0.174 &14.01 & 19.11\\
			Google NIC \cite{vinyals2015show}  & 0.665 & 0.463 & 0.334 & 0.247 & - & - \\
			MS Research \cite{fang2015captions} & - &- & -& -& 20.71 & - \\
			BRNN \cite{karpathy2015deep} &0.646 & 0.455 & 0.303 & 0.201 &- & - \\
			Hard-Attention\cite{xu2015show}         & 0.718 & 0.504 & 0.357 & 0.250 & - & 23.04 \\
			Semantic attention \cite{you2016image} & 0.709 &0.537 & 0.402& 0.304 & - & 0.243 \\
			Joint model with ImageNet \cite{pu2016variational}            & 0.72 & 0.52 & 0.37 & 0.28 & 11.14 & 24.01 \\
			Attributes-CNN+LSTM \cite{wu2015value}  & 0.74 & 0.56& 0.42& 0.31 & 10.49 & 0.26  \\
			RIC with STL  & 0.719 & 0.513 & 0.361 & 0.266 & 11.26 & 23.54 \\
			RIC with STL and DAL  & 0.721 & 0.521 & 0.364 & 0.273 & 11.33 & 23.77 \\
			RIC with variational autoencoder   & 0.734 & 0.535 & 0.385 & 0.299 & 11.41 & 25.43 \\ \bottomrule
		\end{tabular}
	\end{table*}

	\subsection{Data and Settings}
	We employ three image captioning benchmarks for evaluation, including Flickr8k~\cite{hodosh2013framing}, Flickr30k~\cite{young2014image} and Microsoft COCO~\cite{lin2014microsoft} dataset.
	
	\noindent\textbf{Flickr 8k and Flickr 30k~\cite{rashtchian2010collecting}} datasets consists of 8,000 and 31,783 Flickr images, respectively. Most of the images depict humans participating in various activities. Each image is also paired with 5 sentences. Both datasets have a standard training, validation, and testing splits.
	
	\noindent\textbf{MS COCO~\cite{chen2015microsoft,lin2014microsoft}} has 82,783 images and 40,504 validation images, among which some have references in excess of 5. The images are collected from Flickr by searching for common object categories, and typically contain multiple objects with significant contextual information. We apply basic tokenization to MS COCO so that it is consistent with the tokenization presence in Flickr8k and Flickr30k. Specifically, we remove all the non-alphabetic characters in the captions, transform all letters to lowercase, and tokenize the captions using white space. We replace all words occurring less than 5 times with an unknown token <UNK> and obtain a vocabulary of 9,520 words.
	
	For fair comparison, we use the same pre-defined splits for all the datasets as in \cite{karpathy2015deep} and \cite{karpathy2014deep}. We use 1000 images for validation, 1000 for test and the rest for training on Flickr8k and Flickr30k. For MS COCO, 5000 images are used for both validation and testing.
	
	GoogLeNet~\cite{szegedy2015going} or Oxford VGG~\cite{simonyan2014very} are both applicable and can give a boost in performance over the AlexNet~\cite{krizhevsky2012imagenet}. In our implementation, we choose to use VGG-16\cite{simonyan2014very} for ease of comparison. Another important detail is that we use the predefined splits of Flickr8k. However, one challenge for the Flickr30k and COCO datasets is the lack of standardized splits. As a result, we report with the publicly available splits. For all of the three Flickr8k , Flickr30k/MS COCO dataset we used Adam algorithm\cite{kingma2014adam} for optimization.
	
	\subsection{Results}
	
	\textbf{Baseline: Recurrent neural network based language model}
	This is the basic RNN language model developed by \cite{mikolov2010recurrent} , which has no input visual features.\\
	
	\textbf{Baseline: RNN with Oxford VGG-Net Features (RNN+VGG)}
	In place of the BVLC reference Net features, we have also experimented with Oxford VGG-Net \cite{simonyan2014very} features. Many recent papers \cite{mao2014explain} have reported better performance with this representation. We again used the last-but-one layer after ReLU to feed into the RNN model \\
	
	We also compare three different implementation of our architecture:  solely use spatial transformer layer, use both spatial transformer layer and differentiable attention filter, and with variational autoencoder.
	
	We evaluate the quality of the generated sentences by using perplexity, BLEU \cite{papineni2002bleu} , METEOR \cite{banerjee2005meteor} using the COCO captionevaluation tool \cite{chen2015microsoft} .
	Perplexity measures the likelihood of generating the testing sentence based on the number of bits it would take to encode it. The lower the value the better.
	BLEU and METEOR were originally designed for automatic  machine translation where they rate the quality of a translated sentences given several references sentences.
	For BLEU, we took the geometric mean of the scores from 1-gram to 4-gram, and used the ground truth length closest to the generated sentence to penalize brevity.
	For METEOR, we used the latest version. For BLEU, METEOR higher scores are better.

\begin{figure*}[t]
		\begin{center}
			\includegraphics[width=0.8\linewidth]{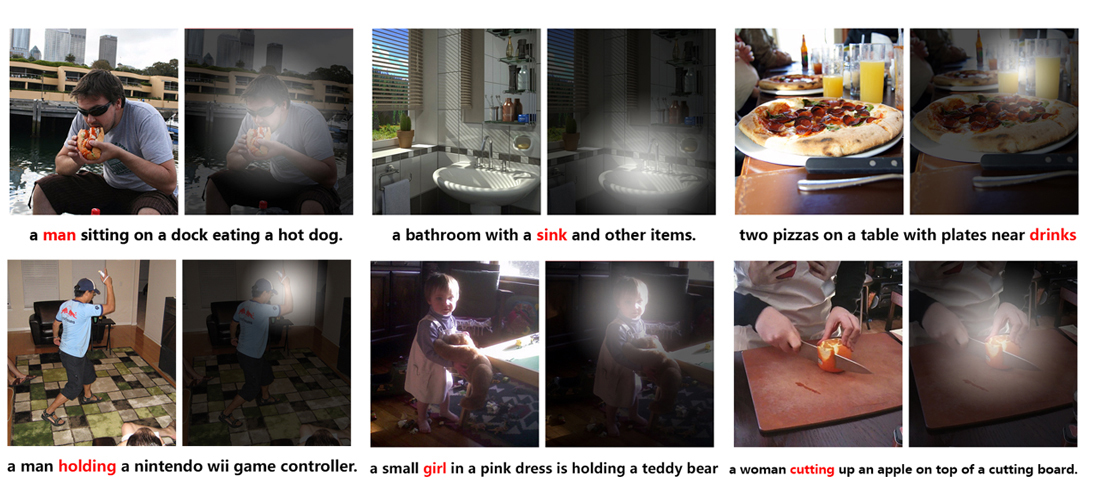}
		\end{center}
		\caption{An example from our method generated result.}
		\label{fig.res2}
	\end{figure*}

\begin{figure*}[t]
		\begin{center}
			\includegraphics[width=0.8\linewidth]{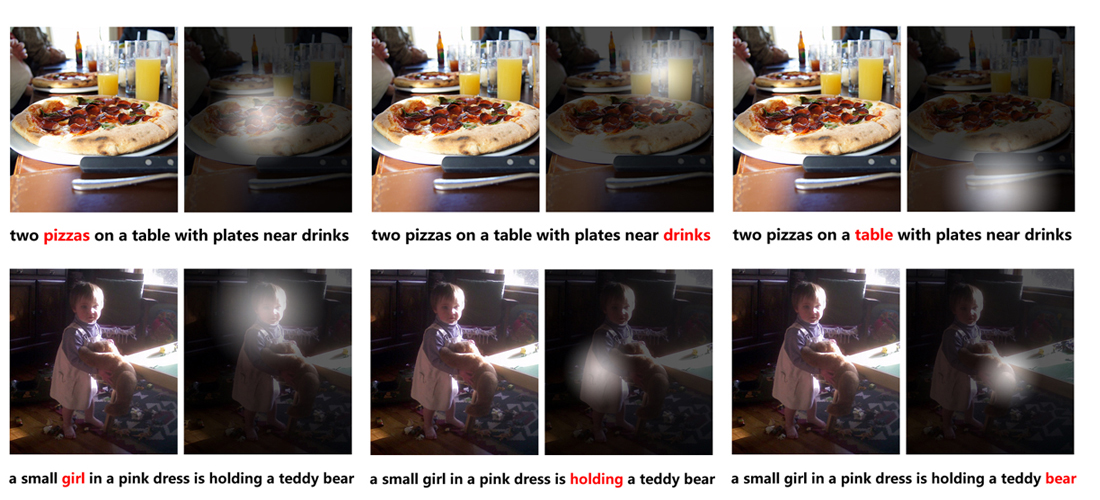}
		\end{center}
		\caption{An example procedure from our method generated result.}
		\label{fig.res1}
	\end{figure*}

	Results for our proposed architecture are reported in Table ~\ref{8k}, ~\ref{30k}, and ~\ref{MSCOCO}. From the tables, it is worth noting that the three different kinds of our model all performs better than most image captioning systems. The basic RIC with STL outperforms almost all previous attention based model such as \cite{xu2015show} and baselines, and the RIC with both STL and DAF improves the results further over basic RIC with STL model, achieves better results over not only attention based method \cite{xu2015show} but also models which include additional information such as semantic attention model in \cite{you2016image}. It is worth noting that by incorporating variational  autoencoder with our proposed architecture to define a novel new loss function, we further improve the results. The only  two methods with better performance than our RIC with both STL and DAF are \cite{you2016image}  and \cite{wu2015value}, the joint model with ImageNet proposed in \cite{you2016image} was training on ImageNet2012 in a semi-supervised manner, the model in \cite{wu2015value} employs an intermediate image-to-attributes layer, that requires determining an extra attribute vocabulary, while our model do not need this additional information. The above analysis of experiment results shows the benefit of our architecture. Examples of generated captions from the validation set of MSCOCO uses the training set of MSCOCO, are shown in Figure ~\ref{fig.res2} ~\ref{fig.res1}.
	
	\section{Conclusion}
	This paper introduced a deep recurrent attention based approach, Recurrent Image Caption(RIC) model, that gives state of the art performance on three benchmark datasets using the BLEU and METEOR metric.  The RIC model which is recurrent model consists of a encoder that sequential and gradually select finer input for the decoder and a decoder that is aimed to either directly maximize caption generation likelihood or maximize a variational lower bound.  When working like the previous,  our method is similar to that of \cite{xu2015show} but a key difference is that our encoder is updated by decoder instead of simply depend on the input features, means that it can capture more fine details about images to generate more accurate captions, as shown in our experiment results. While when working like the later one, our method becomes a deep variational autoencoder in which the autoencoder try to reconstruct input image in the recurrent process and the process guided caption generation In this view, our work is related to \cite{pu2016variational} but a key difference is that in our encoder part, latent code is generated by conditioning on both image and sentence instead of simply use images to encode.  In experiment we show that the above two kinds of  recurrent model achieves state-of-the-art performance on several benchmarks.
	
	% \begin{figure}[t]
	% \begin{center}
	%    \includegraphics[width=0.8\linewidth]{show-eg.PNG}
	% \end{center}
	%     \caption{An example of generated caption from \cite{xu2015show}}
	% \label{fig:example}
	% \end{figure}
	
	% \begin{figure}[t]
	% \begin{center}
	%    \includegraphics[width=0.8\linewidth]{attend.jpg}
	% \end{center}
	%     \caption{An example of generated caption using RIC with raw image as input}
	% \label{fig:example}
	% \end{figure}

	{\small
		\bibliographystyle{ieee}
		\bibliography{temp_egbib}
	}
	
\end{document}